\documentclass{article}

\usepackage{arxiv}

\usepackage[utf8]{inputenc} 
\usepackage[T1]{fontenc}    
\usepackage{hyperref}       
\usepackage{url}            
\usepackage{booktabs}       
\usepackage{amsfonts}       
\usepackage{nicefrac}       
\usepackage{microtype}      
\usepackage{lipsum}
\usepackage{graphicx}
\usepackage{subcaption}
\usepackage{float}

\usepackage{amsmath,amsfonts,amssymb,amsthm}

\title{Stable Lifelong Learning: Spiking neurons as a solution to instability in plastic neural networks}

\author{
 Samuel Schmidgall\\
  U.S. Naval Research Laboratory
  \and \textbf{Joe Hays} \\
 U.S. Naval Research Laboratory \\
}

\begin{document}
\maketitle

\begin{abstract}

Synaptic plasticity poses itself as a powerful method of self-regulated unsupervised learning in neural networks. A recent resurgence of interest has developed in utilizing Artificial Neural Networks (ANNs) together with synaptic plasticity for intra-lifetime learning. Plasticity has been shown to improve the learning capabilities of these networks in generalizing to novel environmental circumstances. However, the long-term stability of these trained networks has yet to be examined. This work demonstrates that utilizing plasticity together with ANNs leads to instability beyond the pre-specified lifespan used during training. This instability can lead to the dramatic decline of reward seeking behavior, or quickly lead to reaching environment terminal states. This behavior is shown to hold consistent for several plasticity rules on two different environments across many training time-horizons: a cart-pole balancing problem and a quadrupedal locomotion problem. We present a solution to this instability through the use of spiking neurons.



\end{abstract}

\section{Introduction}

The extraordinary behaviors produced by biological neural networks are primarily enabled by the persistent intra-lifetime modifications that occur based on accumulated experience. These modifications, called neural plasticity, underlie the foundations of memory, behavioral adaptation, and learning \cite{plasticity_and_memory, Liu2012, annurev.physiol.64.092501.114547, tam2020multiple, neves2008synaptic}. While there are many types of neural plasticity, including structural, connective, and homeostatic, synaptic plasticity has played a particularly central role \cite{itri2008synaptic}. Synaptic plasticity describes when changes in efficacy occur on the connections between neuronal axons and dendrites -- the synapse. These changes develop through the interaction between synaptically connected neurons together with chemical signals diffused throughout the brain. This process is thought to underlie the capacity for both learning and incorporating experience into persistent memories \cite{persistent, itri2008synaptic}.

In the field of Artificial Intelligence (AI), the primary focus of research with ANNs has been on discovering static solutions, where the synaptic weights remain constant throughout the lifetime of the organism. Inspired by the biological brain, a rich history of work has demonstrated the design of ANNs together with synaptic plasticity \cite{soltoggio2018born, miconi2018differentiable, miconi2020backpropamine, najarro2020meta, schmidgall2020adaptive, soltoggio2007evolving}, referred to as Plastic Artificial Neural Networks (PANNs). These networks have shown impressive capabilities in their ability to generalize to novel environmental circumstances, recover from limb damage, enhance memory \cite{najarro2020meta, 10.3389/fnbot.2021.629210, neves2008synaptic}. However, the methods used in these works are trained and operated across finite time horizons. These time horizons generally span a matter of seconds in simulation time, and the ability for these plastic ANNs to generalize beyond the trained time horizon has yet to be examined.

In this work we show that several variants of plasticity in ANNs have difficulty generalizing beyond their initial training time horizon on both tasks of stability and in maintaining reward-driven behavior. Trained under the same conditions, plastic SNNs are shown to generalize to indefinite timespans, demonstrating that SNNs are significantly more stable in the presence of plasticity. This stability is shown to hold across multiple plasticity rules on the same tasks that the plastic ANN failed to generalize to.

\section{Background}

\subsection{Synaptic Plasticity}

A set of learning rules are introduced which control the modification of weights through local activity between neurons. These synaptic learning rules are often designed through evolutionary algorithms, genetic algorithms, and more recently, backpropagation \cite{miconi2018differentiable, miconi2020backpropamine, soltoggio2018born, jordan2020evolving}. A particularly influential class of synaptic learning rules are Hebbian learning rules, which are guided by the statement: "neurons that fire together wire together." This theory claims that an increase in synaptic strength emerges from repeated interaction between pre- and post-synaptic neurons. Hebbian learning theory poses as a framework for explaining \textit{associative} learning, as well as a basis for learning without feedback or reinforcement signals, which is known as unsupervised learning \cite{munakata2004hebbian, pennartz1997reinforcement}. 


Here we consider the general form of Hebbian plasticity. Let $N_{(l)} \in \mathbb{R}$ equal the number of neurons in layer $(l) \in \mathbb{R}$, $W^{(l)}(t) \in \mathbb{R}^{N_{(l)} \times N_{(l-1)}}$ represent the synaptic weights between layer $(l)$ and $(l-1)$ at discrete time $t \in \mathbb{N}$, and $x^{(l)}(t) \in \mathbb{R}^{N_{(l)} \times 1}$ the post-synaptic activity of layer $(l)$ where $x^{(0)}(t)$ is the environment observation-input at time $t$. 

We define the forward pass of stimuli $x^{(l-1)}(t)$ at layer $(l-1)$ through the synaptic weights $W^{(l)}(t)$ as follows,

\begin{equation}\label{Feed Forward}
    x^{(l)}(t) = \sigma(W^{(l)}(t) \times x^{(l-1)}(t)),
\end{equation}

for an arbitrary non-linear function $\sigma(\cdot)$. Additionally, we define the synaptic weight update as,

\begin{equation}\label{Synaptic Update}
    W^{(l)}(t+\delta\tau) = W^{(l)}(t) + \Delta(L, x^{(l-1)}, x^{(l)}).
\end{equation}

Here, $\Delta(\cdot)$ describes a local synaptic update function using pre-synaptic input $x^{(l-1)}(t)$, post-synaptic output $x^{(l)}(t)$, and a set of update parameters $L$.
\\

\textbf{Linear Plasticity}

Among the simplest plasticity rules is the linear Hebbian plasticity rule. With this rule, the co-activation between synapses and weight changes share the following relationship:  

\begin{equation}\label{Hebb Rule W}
    W^{(l)}(t+\delta\tau) = (\textbf{1}-\alpha^{(l)}) \odot W^{(l)}(t) + \alpha^{(l)} \odot (x^{(l)}(t)^\intercal \times x^{(l-1)}(t))
\end{equation}

In this equation, the pre- and post-synaptic activity, $x^{(l-1)}(t)$ and $x^{(l)}(t)$, are multiplied resulting in a co-activation matrix, $x^{(l)}(t)^\intercal \times x^{(l-1)}(t) \in \mathbb{R}^{N_{(l)} \times N_{(l-1)}}$. This matrix is then element-wise multiplied by a learning rate matrix, $\alpha^{(l)} \in \mathbb{R}^{N_{(l)} \times N_{(l-1)}}$, which serves to regulate the rate with which weights are updated. The co-activation matrix follows from Hebbian theory in that large co-activations, which correspond to high-activity between neurons, also result in larger weight changes between them.

\textbf{Oja's Rule}

Inheriting the principle that co-activation produces synaptic change from Hebbian learning, Oja's rule provides built-in stability and correlative capabilities by balancing synaptic potentiation and depression \cite{oja1982simplified}. The primary principle behind Oja's rule is that "forgetting" occurs proportional to both the weight value and the square of the post-synaptic activity.


Oja's rule is defined by the following equation:

\begin{equation}\label{eq:DPSNNOJA}
W^{(l)}(t+\delta\tau) = (\textbf{1}-\alpha^{(l)}) \odot W^{(l)}(t) + \alpha^{(l)} \odot (x^{(l-1)} - W(t)^{(l)} \times x^{(l)})(t)^{\intercal}  \times x^{(l)}(t).
\vspace{1.3mm}
\end{equation}

The role of $\alpha^{(l)}$ is the same as in Equation \ref{Hebb Rule W}, and serves as learning rate for each synapse. Differing however, on the right-hand side of the equation $W(t)^{(l)}$ is multiplied by the post-synaptic activity, $x^{(l-1)}$, which is subtracted from the pre-synaptic trace $x^{(l-1)}$.  This quantity is then multiplied again by the post-synaptic activity, $x^{(l)}$. This modification penalizes unbounded growth, and acts as a regulatory mechanism for the synaptic weights.


\textbf{ABCD Plasticity}

While the simple Hebbian and Oja's plasticity rules are effective models of synaptic plasticity, often greater flexibility in learning representation is desirable. ABCD plasticity enables greater representation by combining learned terms for neural co-activations, pre-synaptic activity, and post-synaptic activity \cite{quadruped_meta}.

The dynamics of ABCD plasticity is as follows:
\begin{equation}\label{ABCD Rule W}
    W^{(l)}(t+\delta\tau) = W^{(l)}(t) + \alpha^{(l)}_{w} \odot \Delta_{ABCD}(t)
\end{equation}

\begin{equation}\label{ABCD Rule Delta}
    \Delta_{ABCD}(t) =(A^{(l)}_{w} + B^{(l)}_{w} + C^{(l)}_{w} + D_{w}^{(l)})(t)
\end{equation}

\begin{equation}\label{ABCD Rule A}
    A^{(l)}_{w}(t) = A^{(l)} \odot (x^{(l)}(t)^\intercal \times x^{(l-1)}(t))
\end{equation}

\begin{equation}\label{ABCD Rule B}
    B^{(l)}_{w}(t) = B^{(l)} \odot (x^{(l)}(t)^\intercal \times \textbf{1}_{(l-1)})
\end{equation}

\begin{equation}\label{ABCD Rule C}
    C^{(l)}_{w}(t) = C^{(l)} \odot (\textbf{1}_{(l)}^\intercal \times x^{(l-1)}(t))
\end{equation}

\begin{equation}\label{ABCD Rule D}
    D^{(l)}_{w}(t) = D^{(l)}_{w}
\end{equation}\,

In these equations, each synaptic weight in layer $l$, $W^{(l)}_{i,j}$, has five corresponding plasticity parameters, $A^{(l)}_{i, j}$, $B^{(l)}_{i, j}$, $C^{(l)}_{i, j}$, $D^{(l)}_{i, j}$, $\alpha^{(l)}_{i, j}$, resulting in an evolved set of five $\mathbb{R}^{N_{(l)} \times N_{(l-1)}}$ matrices which govern the dynamics of the network's intra-lifetime learning. $A_{w}^{(l)}$ acts as the learned correlation term, $B_{w}^{(l)}$ the pre-synaptic term, $C_{w}^{(l)}$ the post-synaptic term, $D_{w}^{(l)}$ as the bias term, and $\alpha_{w}^{(l)}$ as the learning rate. In addition, the initial value of the synaptic weights, $W^{(l)}(t=0)$, are learned. Other initial weight conditions, such as random weights, may be used, however co-evolving the plasticity terms and weights enables the evolutionary dynamics to develop either primarily static-weight or highly-plastic networks which exhibits the relative importance of the learned plasticity. 

Notice that Equation \ref{ABCD Rule A} is the same as the right-hand side of the simple Hebbian rule in Equation \ref{Hebb Rule W}. In this way, ABCD is capable of encapsulating the dynamics of the linear Hebbian rule with additional complexity.

\textbf{STDP}

Perhaps one of the most successful methods for describing the plasticity occurring in the brain is Spike-Timing Dependent Plasticity (STDP) \cite{markram1997regulation, bi1998synaptic}. In the brain, neuronal spikes are often precisely timed in relation to certain events, which is one of the most attractive aspects that STDP is able to capture in learning. A synapse which models STDP is modified in relationship with the precise spike timing of the pre- and post-synaptic neurons. 

Presented in the following equation is a simple variant of STDP:

\begin{equation}\label{stdp sum}
    \Delta W_{j}^{(l)} = \sum_{pre=1}^{N} \sum_{post=1}^{N} \textit{P}(t^{post}_{i} - t^{pre}_{j})
\end{equation}

\begin{equation}\label{stdp P}
    \textit{P}(x) = \begin{cases}
                                   \textit{A}_{+} \text{exp}(-x/\tau_{+}) & \text{for $x \geq 0$} \\
                                   -\textit{A}_{-} \text{exp}(x/\tau_{-}) & \text{if \text{for $x<0$}}
  \end{cases}
\end{equation}

In this equation, $\Delta W_{j}^{(l)}$ is determined by the relationship between the pre- and post-synaptic firing times, $t^{post}_{i} \in \mathbb{R}$ $t^{pre}_{j} \in \mathbb{R}$ $P(x)$ across all firing times that occurred $N \in \mathbb{N}$. $P(x)$ in Equation \ref{stdp P} denotes the STDP function, which takes the distance between spike times, $t^{post}_{i} - t^{pre}_{j}$, and returns exponentially decaying values scaled by $A_{+,-} \in \mathbb{R}$ and $\tau_{+,-} \in \mathbb{R}$ based on whether the post-spike arrives after the pre-spike, $x \geq 0$, or the pre- before post-spike, $x<0$. One issue with this model, however, is that iterating through all previous spike times for new activity is expensive. In practice, this model can be implemented as an online rule using synaptic traces:

\begin{equation}
    \tau_{+} \frac{dx}{dt} = -x_{j} + a_{+}(x_{j})\sum_{pre}\delta(t-t^{pre}_{j})
\end{equation}

\begin{equation}
    \tau_{-} \frac{dy}{dt} = -y_{j} + a_{-}(y)\sum_{post}\delta(t-t^{post})
\end{equation}

\begin{equation}
    \Delta W_{j}^{(l)} = \textit{A}_{+} (W_{j}) x(t) \sum \delta (t-t^{n}) - \textit{A}_{-} (W_{j}) y(t) \sum \delta (t-t^{f}_{j})
\end{equation}

Here, the role of each variable is the same as in Equations \ref{stdp sum}-\ref{stdp P}. The primary difference is that synaptic traces are introduced, x and y respectively, which accumulate an exponentially decaying history of the synapse spiking activity.

\subsection{Spiking Neurons}

Spiking neural networks more closely resemble biological neural networks in that information is transmitted asynchronously through sparse binary signals \cite{10.3389/fnins.2018.00774}. In the biological brain, these signals are emitted through electrical "spikes" down axonal connections to connected neurons. These electrical spikes are integrated into the neuron's electrical storage system as membrane potential, and once the membrane potential reaches a critical threshold, it emits its own spike. While the magnitude of these spikes are largely identical across neurons, variation in signal strength, whether producing potentitation or depression, is a product of the size and strength of the synaptic terminal connecting neurons \cite{sudhof2018towards}. This is largely where the idea of "weights" in artificial and spiking neural network arise from. Like the biological neuron, in addition to emitting binary spikes, spiking neurons also have internal membrane potential which is used to decide whether or not to send a spike. Once the membrane reaches a particular threshold, a spike is emitted, and the membrane potential resets. This behavior can be modelled in various ways, several of which are presented below.


\noindent
\textbf{Integrate-and-Fire}

One of the earliest, and arguably the simplest, neuron models is the integrate-and-fire, which is also referred to as the perfect integrate-and-fire, or the non-leaky integrate-and-fire. In this model, incoming activity is integrated and stored over time, and a binary-valued signal of one is emitted when the membrane potential is greater than a given threshold $V_{th} \in \mathbb{R}$, and zero otherwise. This neuron model assumes perfect insulation in that membrane potential does not passively return to a fixed-point over time. The membrane potential dynamics are as follows:
\begin{equation}
    \tau_{m} \frac{dV_{m}(t)}{dt} = R_{m} I(t),
\end{equation}
where $V_{m}(t) \in \mathbb{R}$ is the membrane potential at time \textit{t}, $\tau_{m} \in \mathbb{R}$ is the membrane time constant, $R_{m} \in \mathbb{R}$ is the current resistance, and $I(t) \in \mathbb{R}$ is the input current at time \textit{t}.

Once the membrane potential crosses its spiking threshold $V_{th} \in \mathbb{R}$ the neuron emits a spike and the membrane potential resets to a baseline value, or resting potential, which is typically set to zero.\\

\noindent
\textbf{Leaky Integrate-and-Fire}

The Leaky Integrate-and-Fire (LIF) neuron model is derived from the integrate-and-fire model, however, with the addition of a decay, or "leak," component on the membrane potential. The leak slowly over time brings the membrane potential back to its resting equilibrium. In terms of biology, this mechanism is a product of passive ion diffusion through the neuron membrane which does not act as a perfect source of insulation. This passive diffusion is thought to add beneficial dynamic complexity for learning.

The LIF membrane potential dynamics can be modelled as follows:
\begin{equation}
    \tau_{m} \frac{dV_{m}(t)}{dt} = R_{m}I(t) - [V_{m}(t) - v_{rest}],
\end{equation}
where $V_{m}(t)$ is the membrane potential at time \textit{t}, $\tau_{m}$ is the membrane time constant, $R_{m}$ is the current resistance, $v_{rest} \in \mathbb{R}$ is the membrane resting potential, and $I(t)$ is the input current at time \textit{t}. Like the integrate-and-fire, the LIF emits a spike and the membrane potential resets to $v_{rest}$ once a fixed-valued potential threshold $V_{th}$ is surpassed.

One disadvantage of the neuron model's simplicity is that it does contain adaptive parameters that change the neuron's behavior over time, and thus can not accurately describe "in-vivo" experimentally measured spiking activity \cite{fuortes1962interpretation}. Despite this, the LIF neuron model remains the most utilized model in both learning experiments and neuromorphic hardware implementations likely due to its balance between computational simplicity and dynamic complexity.\\





\subsection{Evolutionary Strategies}

The initial parameter vector of size M $\in \mathbb{R}$, $\theta_{t} \in \mathbb{R}^{M}$, represents values which are to be modified across generations on an evolutionary timescale, as opposed to intra-lifetime modifications. To begin the evolutionary cycle, $\theta_{t}$ is either randomly initialized or set to zero and a population of N random noise vectors, $v_{i, t} \in \mathbb{R}^{M}$, are sampled from a random distribution with a standard deviation $\sigma \in \mathbb{R}$ to create a population of individuals $p_{i, t} = \theta_{t} + v_{i, t}$, where $i$ indexes an individual in the population, and $t \in \mathbb{N}$, the evolutionary time-step. Each individual is then evaluated over the course of a lifetime using an environment defined reward, $r_{i, t}  \in \mathbb{R}$. Using the corresponding information, and a learning rate $\alpha \in \mathbb{R}$, parameters are updated as follows:

\begin{equation}
    \theta_{t+1} = \theta_{t} + \alpha \frac{1}{N\sigma^{2}}\sum^{N}_{n=1} v_{i, t}r_{i,t}
\end{equation}

There are numerous advantages in using evolutionary optimization over alternative methods. Due to the independent nature of individual evaluation, evolutionary methods are well suited for distributed computation enabling a much shorter training time in comparison to backpropagation-based methods. In addition, these methods, as a product of not using backpropagated gradients, do not perform backpropagation through time, and hence time-dependent parameters, like synaptic plasticity, do not require immense compute time. Particularly in the context of synaptic plasticity, some situations display evolutionary algorithms outperforming gradient-based approaches in both learned performance and in training time \cite{schmidgall_adaptive, salimans2017evolution}.

The lifetime of an individual, along with the evaluation reward, is heuristically defined. Since creating the next population of individuals \textit{requires} the parent population to have finished evaluation, and hence their lifetime, often the lifetime of each individual is hard-limited to a truncated lifespan called the "time-horizon."

Evolutionary strategies methods have the advantage of being a "black box" optimization method, meaning optimization occurs without higher-order information, only observing the function input and output pairs. This is particularly desirable in the setting of Reinforcement Learning, as its black box nature enables it to be invariant to action and reward frequency, tolerant of long time horizons, and not requiring the need for value function approximation \cite{salimans2017evolution}.

\section{Experiments and Results}


\subsection{Simple Balancing Problem}

\begin{center}
    \begin{figure}[H]
        \centering
        \includegraphics[width=.48\linewidth]{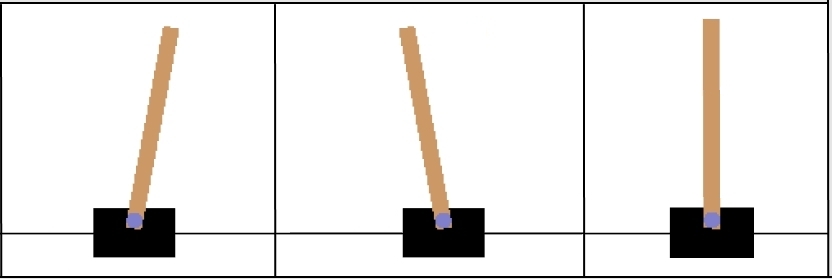}\\
        \caption{Depiction of the Cart-Pole balancing task in three scenarios. (Left) The pole is falling to the right, and hence a force must be applied rightward to the cart on the left side to counteract the falling. (Center) The pole is falling to the left, and hence a leftward force must be applied. (Right) The pole is balanced, requiring no force application for stability.}
        \label{fig:CartPole}
    \end{figure}
\end{center}

Among the simplest of control problems is the discrete-action cart-pole balancing task (Figure \ref{fig:CartPole}). In this problem, a pole is situated at the center of a cart which lies on a track. Force may be applied to the cart left or right, and the objective is for the pole to retain an upward position. If the pole falls beyond a certain angle, the simulation is terminated, and the "lifespan" of the task is cut short. Additionally, to prevent the simulation from running indefinitely, a maximum lifespan is specified which is deemed as the time horizon. The time horizon is necessary for most modern learning schemes, since comparison between multiple lifetimes, often called rollouts, is necessary for the algorithm to update parameters. The time horizon is of great importance especially for dynamical models because it determines how long the model is expected to perform a given task, and the learned parameters are optimized specifically for this time duration.


While the Cart-Pole task does not necessarily require intra-lifetime learning, and can be solved without adaptation, it is precisely for this reason that it serves as a quality example. Problems necessitating lifelong learning often require both stability and adaptivity, and ideally, the distinction between such should not require a great deal of manual design. As in, adaptive models, such as plasticity, should provide the capacity for handling both tasks which require adaptivity and those which require stability. Of course most problems belong on a scale for either, however, it is often in the extremes that underlying issues unveil themselves in full honesty.

To examine the effect of time horizon on the long-term stability of plastic neural networks, a collection of models are trained at varying time horizon lengths. These trained models are then evaluated in the same environment without a time horizon upper-bound. Once the pole falls beyond a certain threshold, the environment stops execution, and the duration of balance is recorded and compared with the corresponding time horizon. The trained time horizons span from 100 to 3000 timesteps of 0.02 seconds of integration, totaling 2-60 seconds in real-time; the original benchmark time horizon for the CartPole task is 200 steps. In this experiment plastic ANNs are evaluated on the ABCD rule and Oja's rule and plastic SNNs are evaluated on STDP and Oja's rule. Both feedforward and recurrent neural network topologies are examined across multiple time horizon trainings.

Oja's rule (Equation \ref{eq:DPSNNOJA}) typically operates with real numbered activity, and hence to allow for an adequate comparison between the ANN and SNN with Oja's rule, the SNN Oja's update rule uses spiking activity averaged across several timesteps to produce a rate-based value instead of updating with binary spikes (Equation \ref{eq:SNNOJA}). 

\begin{equation}\label{eq:SNNOJA}
W^{(l)}(t+\Delta\tau) = (1-\eta^{(l)})W^{(l)}(t) + \eta^{(l)}(\frac{\sum^{N}_{i=0} x_{i}^{(l)}(t)}{N} - W(t)^{(l)} \frac{\sum^{N}_{i=0} x_{i}^{(l-1)}(t)}{N})^{\intercal}\frac{\sum^{N}_{i=0} x_{i}^{(l-1)}(t)}{N}.
\vspace{1.3mm}
\end{equation}

For the feedforward networks, there were two fully-connected hidden layers of 32 neurons. The recurrent networks had the same structure, but with recurrent connections on both hidden and the output layers. The hidden layers of the ANNs used a hyperbolic tangent activation function, which maps incoming activity between -1 and 1. The network topology was identical across all CartPole experiments for ANNs and SNNs.

\begin{figure}
    \begin{subfigure}{.5\textwidth}
        \centering
        \includegraphics[width=.95\linewidth]{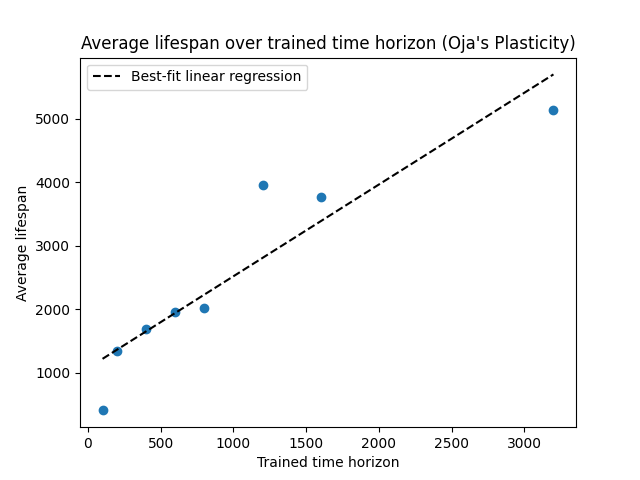}
    \end{subfigure}%
    \begin{subfigure}{.5\textwidth}
        \centering
        \includegraphics[width=.95\linewidth]{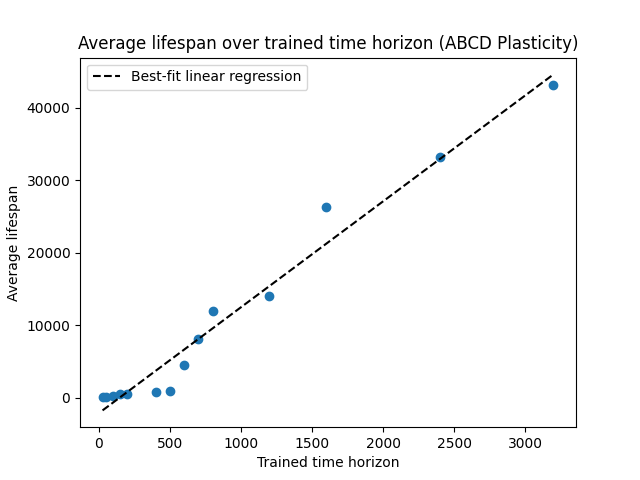}
    \end{subfigure}
    \caption{Linear regression fit on data points from average uncapped lifespan (y-axis) for plastic artificial neural networks (Oja's, left; ABCD, right) based on the time horizon used during training (x-axis) on the CartPole-v1 control environment. With the linear regression equation ($y = Ax + B$), Oja's $A=1.44$ and ABCD $A=14.78$ indicates an expected lifespan of x1.44 and x14.78 beyond the training time horizon with Oja's and ABCD respectively.}
\label{fig:cartpole_feedforward_ann}
\end{figure}

Figure \ref{fig:cartpole_feedforward_ann} displays the average post-training lifespan over the trained time horizon for the ANN with Oja's and ABCD plasticity. Figure \ref{fig:cartpole_feedforward_ann} is fit with a linear regression curve to demonstrate the predicted average lifespan based on the trained time horizon. The regression curve fit for the plastic RNNs are shown in the Appendix. In both the feedforward and recurrent networks, the ABCD rule was more stable, providing a predicted expected lifespan of x14.78 beyond the training time horizon for the feedforward, and x1.29 for the recurrent. Oja's rule only enables a x1.44 and x1.04 for feedforward and recurrent respectively. The recurrent networks were remarkably less stable in both instances for the ANN, and the feedforward networks were not shown to be capable of finding a stable solution even after training on an extended time horizon from the original 200 timesteps to 3000.

In the plastic SNNs, a linear curve does not suffice to model the expected lifespan, as after any training time horizon beyond 400 timesteps the network is capable of balancing indefinitely\footnote{The maximum tested was 10 million timesteps.}. This stable behavior holds for both the plastic feedforward and recurrent SNNs. Additionally, the training time needed to attain an "indefinite" balancing lifespan was consistent for both the STDP and Oja's rule, even when Oja's rule used a rate-based representation of pre- and post-synaptic activity.


\begin{figure}
    \begin{subfigure}{.5\textwidth}
        \centering
        \includegraphics[width=.95\linewidth]{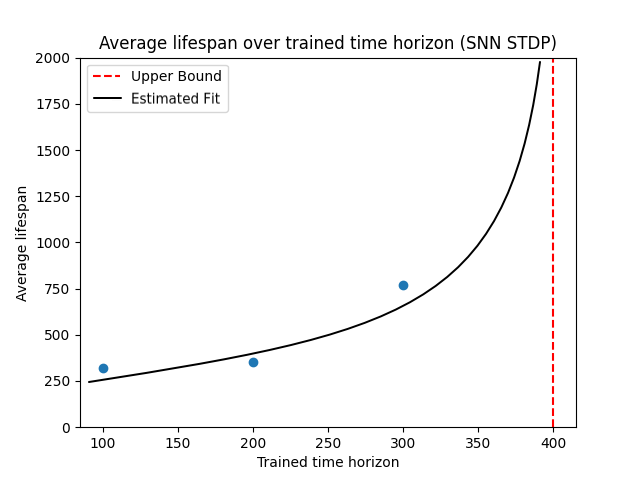}
    \end{subfigure}%
    \begin{subfigure}{.5\textwidth}
        \centering
        \includegraphics[width=.95\linewidth]{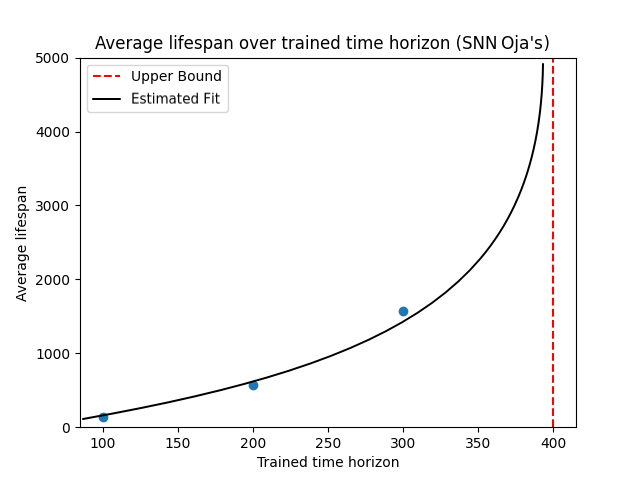}
    \end{subfigure}
    \caption{Estimated fit lines on data points from average uncapped lifespan (y-axis) for plastic spiking neural networks (Oja's, left; STDP, right) based on the time horizon used during training (x-axis) on the CartPole-v1 control environment.}
\label{fig:cartpole_feedforward_snn}
\end{figure}

\subsection{Quadrupedal Locomotion}

\begin{figure}[H]
    \centering
    \includegraphics[width=0.95\linewidth]{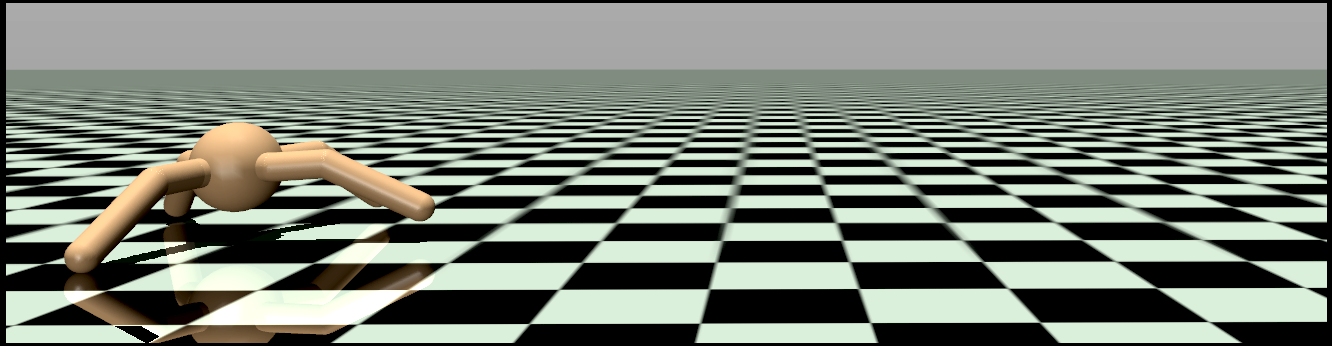}
    \caption{Quadrupedal Locomotion environment.}
    \label{fig:ant}
\end{figure}

In \cite{quadruped_meta}, plastic ANNs were shown to be impressively capable of generalizing beyond what had occurred during training time. In one experiment, several limbs of a quadruped (Figure \ref{fig:ant}) were disabled randomly during training, whereas others were kept functional. The limbs which remained functional during training time served as a "training set," so that several disabled limbs could be examined as a "testing set" after training to determine whether generalization within intra-lifetime learning was actually occurring. Indeed, it was shown that locomotion progressed acceptably on one of the two testing set limbs, with a rapid recovery in performance when disabled mid-locomotion. However, the functionality of the quadruped beyond the initial training time horizon was never considered in their results. This section replicates these experiments without damaged morphology, and observes the long-term stability of these solutions both in terms of qualitative behavior and obtained reward.

In this experiment, the Ant quadrupedal locomotion environment (Figure \ref{fig:ant}) is trained with the same parameters and algorithm\footnote{We use the Mujoco simulator instead of Bullet; however, the environment uses the same robotic body and emulates near-identical physics.} as in \cite{quadruped_meta}. The Ant environment differs from CartPole in that there is no natural definition of "lifespan;" the simulation never officially finishes execution unless a time horizon is instantiated. In \cite{quadruped_meta}, the ability to retain positive reward-driven behaviors was considered to be generalizing across disabled limbs, and hence generalization was measured with reward over time. Considering that, to measure generalization across time, the ability to retain reward-driven behavior beyond the training time horizon is examined.

Much like in the CartPole experiment, this experiment considers the training of plastic networks across several different time horizons; in this case, a time horizon of 250, 500, and 1000. The ABCD-plasticity rule used in \cite{quadruped_meta} is applied to both the ANN and SNN, where, for the SNN, the pre- and post-synaptic activity uses rate-based spiking activity like in Equation \ref{eq:SNNOJA}. 

Figures 6-7 display the accumulated reward over time for a simulation run on the trained ANN (Figure 6) and SNN (Figure 7) on the 250, 500, and 1000 timestep-trained networks. In the ANN scenario, each of the three trained networks no longer continue accumulating positive reward the moment the training time horizon is reached. Beyond the training time horizon in the case of the 250 timestep ANN, reward not only ceases to accumulate, but the quadruped begins obtaining negative reward, where in this scenario, the robot flips onto its back and exerts unnecessary leg movement without making forward progress. The 500 and 1000 timestep cases are similar, with the quadruped quickly obtaining positive reward until the time horizon ends, and the curve flattens out. This indicates that on the task of quadrupedal locomotion, ABCD plasticity struggles to generalize to time in ANNs.

Surprisingly for the SNN, the average performance \textit{within} the time horizon on each training is slightly lower than the ANN. However, generalization to time is significantly better, with there being little degradation in reward accumulation after the end of the trained time horizon. Each of the three networks demonstrate the capability to continue performing locomotion without functional deterioration; however, this generalization is shown to improve with a larger time horizon. In the 250 timestep network training, the reward accumulation slows down over time, but generally continues upward. In the 500 timestep network, it follows a linear upward trend up until around 2000 timesteps, where it follows a slower upward linear trend. Finally, in the 1000 timestep network, as far as it was observed, obtained a linear upward trend in performance, indicating strong generalization to time beyond the training time horizon.

\begin{center}
    \begin{figure}
        \includegraphics[width=0.95\linewidth]{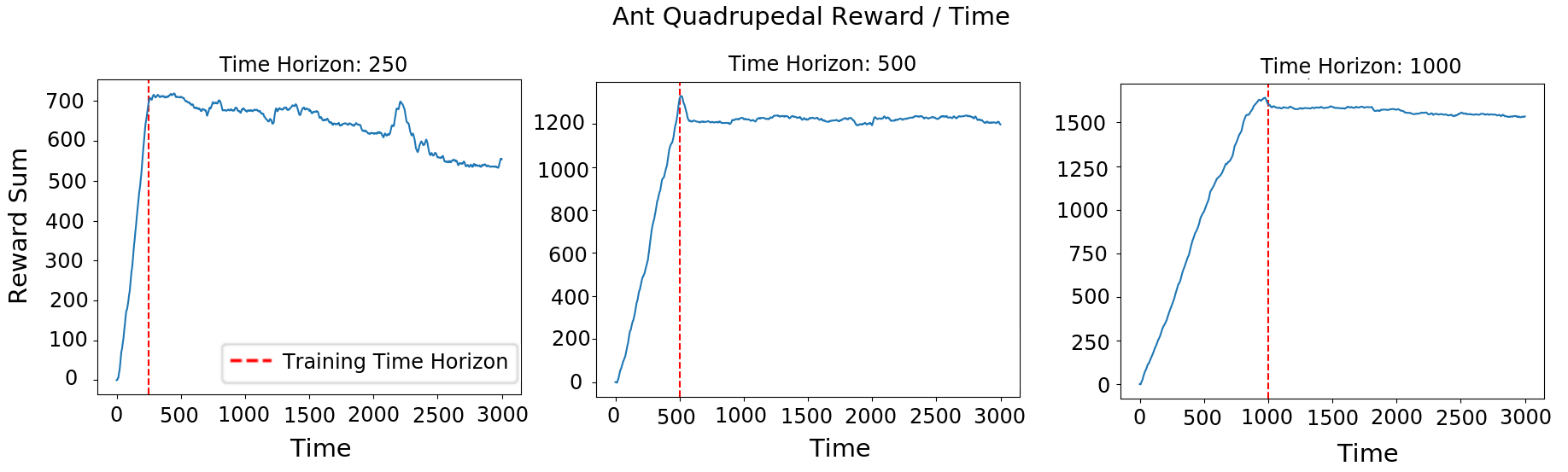}
        \caption{The reward accumulation over time of three ABCD-plastic ANNs trained on Ant quadruped for three different training time horizons. Each of the three networks display near-instantaneous reward decline when evaluated beyond their trained time horizon (vertical red-dashed line).}
    \end{figure}
\end{center}

\begin{center}
    \begin{figure}
        \includegraphics[width=0.95\linewidth]{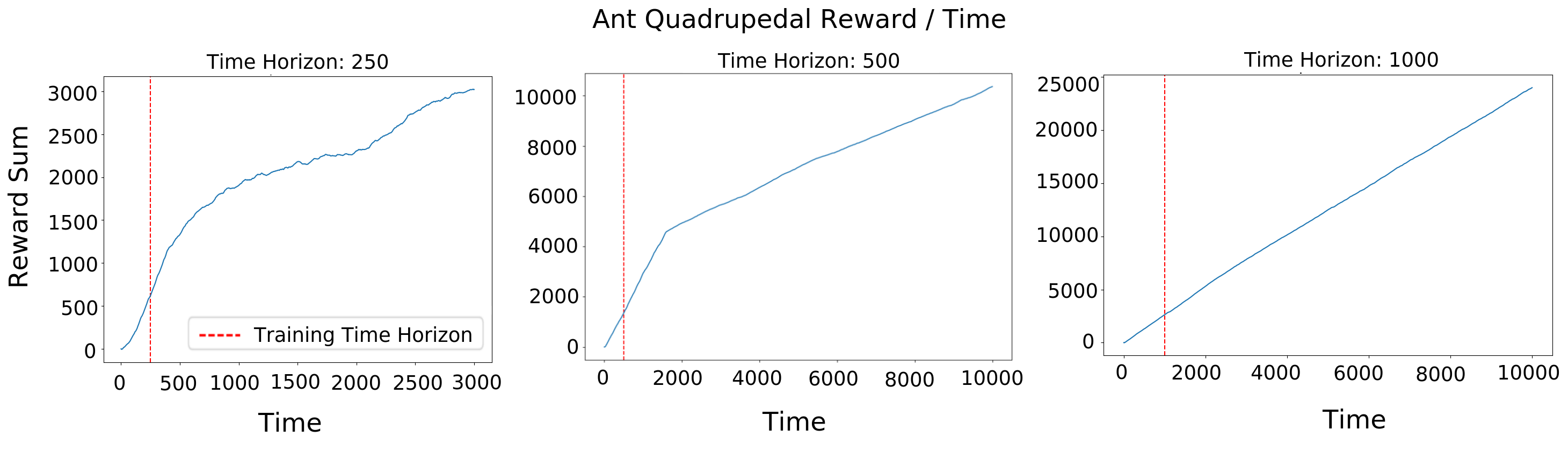}
        \caption{Reward accumulation over time for three ABCD-plastic SNNs trained on Ant quadruped for three different training time horizons. All of the networks retain stable increase in reward beyond the training time horizon, which becomes more stable with a longer horizon.}
    \end{figure}
\end{center}








\section{Discussion}

Synaptic plasticity is a powerful mechanism for unsupervised learning in neural networks, inspired by learning processes in the biological brain \cite{plasticity_and_memory, Liu2012, annurev.physiol.64.092501.114547, tam2020multiple, neves2008synaptic}. This process has been incorporated into spiking and artificial neural networks to enable intra-lifetime learning \cite{soltoggio2018born, miconi2018differentiable, miconi2020backpropamine, najarro2020meta, schmidgall2020adaptive, soltoggio2007evolving}. However, in this work it was shown that plastic ANNs struggle to generalize their behavior beyond the training time horizon. SNNs on the other hand are shown to effectively generalize to time on a cart-pendulum task requiring fine-tuned stability, and on a quadrupedal locomotion task where forward progress is rewarded. In the cart pendulum task, ANNs were evaluated with Oja's plasticity and ABCD plasticity rules, and the training time horizon was shown to have a linear relationship with the the average lifespan. SNNs evaluated with Oja's rule and STDP were shown to have lifespan divergence around a 400 timestep time horizon, where behavior generalized to time and the network was capable of balancing the pole indefinitely. In the quadrupedal locomotion task, plastic ANNs were shown to degrade in performance the moment after the time horizon was reached, whereas SNNs were shown to continue improving in performance.

Spiking neural networks have the additional advantage of possibly being used together with neuromorphic hardware, which increases the energy efficiency of deployed networks by orders of magnitude compared with ANNs \cite{schuman2017survey}. Often the capability for incorporating plasticity can be designed into neuromorphic hardware, which further reinforces the benefits of plastic SNNs for use in application. It will likely be necessary for robotic applications of plastic SNNs to demonstrate generalization to time, as the majority of modern learning algorithms require a finite time horizon which is not representative of the application time that the robotic learner will be used for in the real world. In many scenarios, the target application time will be indefinite, and long-term stability will not only be desirable, but necessary.

The importance of this work is in highlighting potential issues with plastic ANNs, and presents the use of spiking neurons as a solution. The purpose of synaptic plasticity is to allow learning to occur within and beyond the training period of a neural network, and hence it is necessary to consider the ability to generalize not only in the task domain but also time domain. We hope that this work brings to light potential issues with generalization to time in plastic ANNs, and provides a potential solution for this instability. We also hope that this work steers future research with plastic neural networks in the direction of considering time as a domain which is considered for generalization benchmarking. 


\bibliographystyle{unsrt}
\bibliography{references}

\end{document}